\documentclass[sigconf]{acmart}
\AtBeginDocument{%
  }
\usepackage{multirow}
\usepackage{subcaption}
\acmConference[ISMSI '25]{April 26--28,
  2025}{Tokyo, Japan}

\renewcommand\footnotetextcopyrightpermission[1]{}

\begin{document}

\title{Comparative Analysis of Pooling Mechanisms in LLMs: A Sentiment Analysis Perspective}


\author{Jinming Xing}
\affiliation{%
  \institution{North Carolina State University}
  \country{}}
\email{jxing6@ncsu.edu}


\author{Dongwen Luo}
\affiliation{%
  \institution{South China University of Technology}
  \country{}}
\email{976267567ldw@gmail.com}

\author{Chang Xue}
\affiliation{%
  \institution{Yeshiva University}
  \country{}}
\email{cxue@mail.yu.edu}

\author{Ruilin Xing}
\affiliation{%
  \institution{Guangxi University}
  \country{}}
\email{ruilinxing8@gmail.com}

\renewcommand{\shortauthors}{Jinming et al.}

\begin{abstract}
  Large Language Models (LLMs) have revolutionized natural language processing (NLP) by delivering state-of-the-art performance across a variety of tasks. Among these, Transformer-based models like BERT and GPT rely on pooling layers to aggregate token-level embeddings into sentence-level representations. Common pooling mechanisms such as Mean, Max, and Weighted Sum play a pivotal role in this aggregation process. Despite their widespread use, the comparative performance of these strategies on different LLM architectures remains underexplored. To address this gap, this paper investigates the effects of these pooling mechanisms on two prominent LLM families -- BERT and GPT, in the context of sentence-level sentiment analysis. Comprehensive experiments reveal that each pooling mechanism exhibits unique strengths and weaknesses depending on the task's specific requirements. Our findings underline the importance of selecting pooling methods tailored to the demands of particular applications, prompting a re-evaluation of common assumptions regarding pooling operations. By offering actionable insights, this study contributes to the optimization of LLM-based models for downstream tasks.
\end{abstract}

\ccsdesc[500]{Computing methodologies~Machine learning}
\ccsdesc[500]{Information systems~Data mining}
\ccsdesc[500]{Information systems~Information retrieval}
\ccsdesc[500]{Computing methodologies~Natural language processing}
\keywords{Large Language Models, Pooling Mechanisms, Sentiment Analysis, Token Embedding Aggregation}


\maketitle

\section{Introduction}
Large Language Models (LLMs) have emerged as transformative tools in natural language processing (NLP), offering unparalleled performance across a broad spectrum of tasks. Among these, BERT (Bidirectional Encoder Representations from Transformers) \cite{bert18bert} and GPT (Generative Pre-trained Transformer) \cite{gpt18improving} stand out as two of the most influential architectures. BERT's bidirectional attention mechanism enables it to deeply understand contextual relationships within text, making it particularly effective for comprehension-based tasks. On the other hand, GPT's unidirectional design and autoregressive modeling excel in generating coherent and contextually appropriate text. Together, these models exemplify the state of the art in leveraging transformer-based architectures to tackle diverse linguistic challenges.

The applications of LLMs can be broadly categorized into token-level and sentence-level tasks \cite{floridi20gpt}. Token-level tasks, such as named entity recognition and part-of-speech tagging, require the model to process and predict attributes for individual tokens within a sequence. Sentence-level tasks, including sentiment analysis, entail aggregating token-level information to derive an overall understanding of the text's meaning. These tasks depend heavily on effective mechanisms to condense token embeddings into coherent sentence-level representations. In this study, we focus on sentence-level tasks, particularly sentiment analysis, as it serves as a fundamental benchmark for evaluating the semantic capabilities of LLMs.

Pooling layers play a critical role in sentence-level tasks by aggregating token embeddings into unified representations \cite{yao24survey}. Commonly employed pooling strategies include Mean pooling, which averages embeddings to provide a balanced view, Max pooling, which captures the most salient features, and Weighted Sum pooling, which applies learned weights to emphasize contextually significant tokens. Despite their importance, the comparative effects of these pooling mechanisms on different LLM architectures remain underexplored. To bridge this gap, this paper evaluates the performance of these pooling strategies on BERT and GPT within the context of sentiment analysis.
Our contributions are threefold:
\begin{itemize}
    \item We provide a comprehensive evaluation of Mean, Max, and Weighted Sum pooling mechanisms on BERT and GPT models.
    \item We identify task-specific strengths and limitations of each pooling method, offering insights into their optimal use cases.
    \item We present actionable recommendations for practitioners to select appropriate pooling mechanisms based on the specific requirements of downstream tasks.
\end{itemize}

\section{Related Work}
Deep Learning models \cite{cheng24patch,cheng25unifying,cheng22deep,cheng22estimation,xing24traffic,cheng23shapnn}, especially the Transformer-based \cite{yangbotnet,linresearch,yangrac,li24contextual,luo24rmgen} have found innovative applications across various fields like image analysis \cite{zhaoresearch, yangcnn}, virtual reality \cite{yangzhan}, sequences modeling \cite{dengresearch}, medical diagnosis \cite{yang24tcell}, and emotion recognition \cite{yangarxiv1}.

\textbf{The transformer architecture}, proposed by Vaswani et al. \cite{vas17attention}, addressed the sequential computation limitations of RNNs through self-attention mechanisms. This innovation enabled parallel processing of input sequences and more effective modeling of long-range dependencies, establishing the foundation for modern LLMs. The transformer's encoder-decoder architecture demonstrated superior performance in machine translation tasks and quickly became the de facto standard for neural sequence transduction models \cite{xing24enhancing}.

\textbf{BERT}, introduced by Devlin et al. \cite{bert18bert}, marked a breakthrough in NLP by utilizing a bidirectional transformer architecture to generate contextual embeddings for tokens. Unlike unidirectional models, BERT captures relationships between words in both forward and backward directions, enabling superior understanding of text semantics. Numerous variations of BERT have since been developed to enhance its performance and efficiency. For example, RoBERTa \cite{liu19roberta} optimized BERT's training process by removing the next sentence prediction task and training on a larger dataset. DistilBERT \cite{sanh19distil} focuses on reducing model size while retaining most of BERT's performance, making it suitable for resource-constrained environments. BERT and its variants have been widely adopted for tasks such as sentiment analysis, text classification, and question answering.

\textbf{GPT}, initially proposed by OpenAI \cite{gpt18improving}, pioneered autoregressive modeling for text generation. The model predicts the next word in a sequence based on prior context, making it particularly effective for generative tasks such as text completion and summarization. Its successor, GPT-2 \cite{radford19language}, significantly expanded the model's capacity and demonstrated remarkable versatility across tasks without task-specific fine-tuning. The introduction of GPT-3 \cite{brown20language} and GPT-4 \cite{achiam23gpt} further advanced the state of the art, leveraging billions of parameters to perform complex tasks such as language translation and dialogue generation with minimal instruction. GPT models have found extensive applications in content creation, conversational agents, and zero-shot learning scenarios \cite{lund23chatting}.

\textbf{Pooling mechanisms} are critical for aggregating token embeddings into sentence-level representations. Mean pooling, one of the simplest techniques, computes the average of token embeddings, providing a balanced view of the input. Max pooling, in contrast, captures the most salient features by selecting the maximum value along each dimension. Weighted Sum pooling introduces a learnable weighting mechanism to prioritize contextually important tokens \cite{gao22param,xing22weighted}. These pooling strategies have been explored in various contexts \cite{xing22weighted}. For instance, Justyna et al. \cite{sar21detection} incorporated Mean pooling in their work on contextualized embeddings for sentence classification. Max pooling was applied by Conneau et al. \cite{conneau17supervised} in supervised learning tasks to enhance feature selection. Despite these efforts, the comparative impact of pooling operations on LLMs such as BERT and GPT remains underexplored, especially for sentence-level tasks like sentiment analysis.

\section{Methodology}

\subsection{Attention Mechanism Fundamentals}
The attention mechanism serves as the architectural cornerstone of transformer-based Large Language Models (LLMs) \cite{yao24survey}. At its core, attention allows models to dynamically weight the importance of different tokens when processing sequential data. The standard attention mechanism can be mathematically formulated as:
\begin{equation}
    \text{Attention}(Q, K, V) = \text{softmax}\left(\frac{QK^T}{\sqrt{d_k}}\right)V
\end{equation}
where:
\begin{itemize}
    \item $Q$ represents query matrices
    \item $K$ represents key matrices
    \item $V$ represents value matrices
    \item $d_k$ is the dimension of the keys
\end{itemize}

This mechanism enables models to capture complex contextual relationships by allowing each token to attend to all other tokens in a sequence, generating rich, context-aware representations.

\begin{figure}[htbp]
    \centering
    \includegraphics[width=0.5\linewidth]{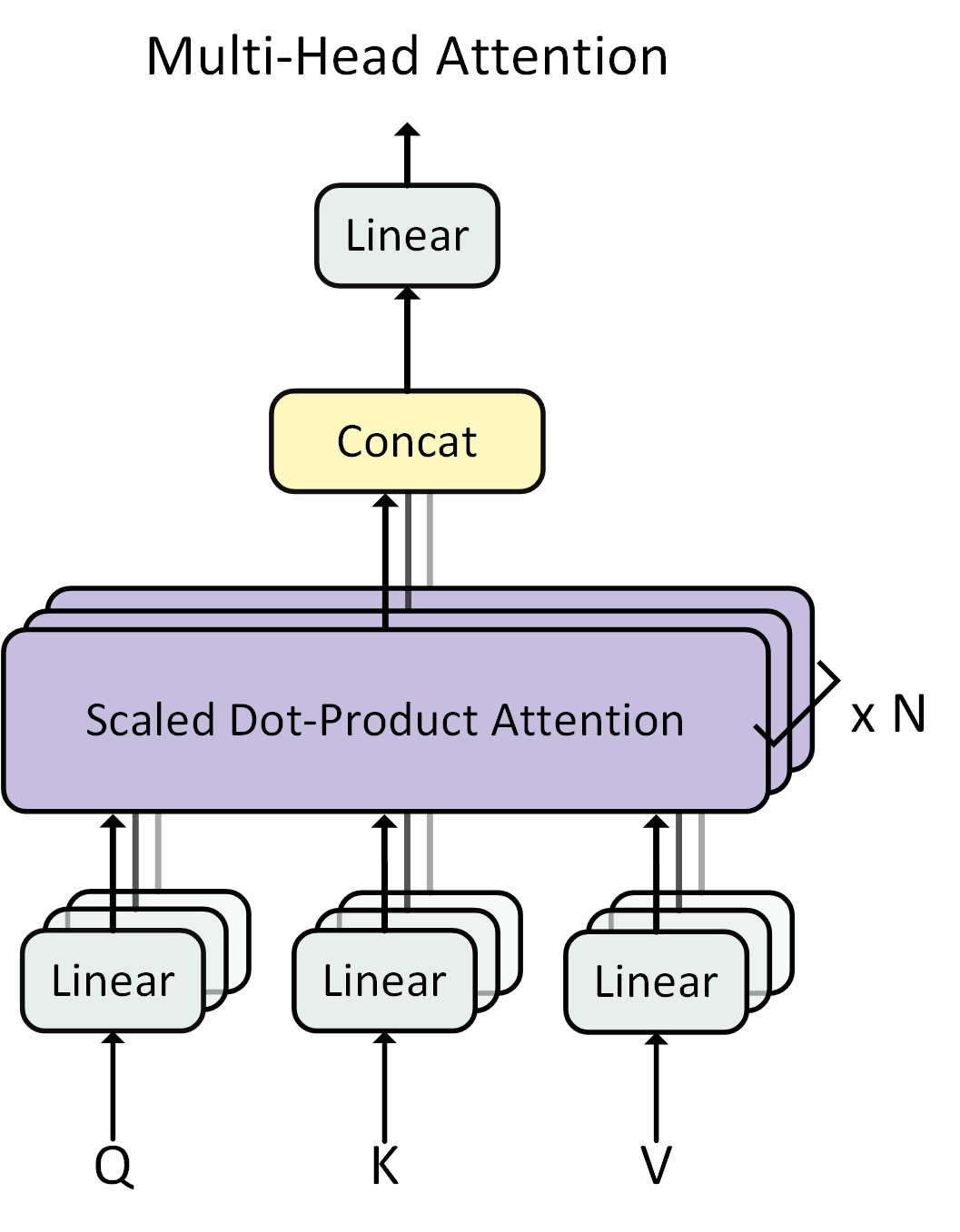}
    \caption{Multi-Head Attention Mechanism}
    \label{fig:multi_head_attention}
\end{figure}

In practice, multi-head attention mechanisms \cite{xing24enhancing}, as shown in Figure \ref{fig:multi_head_attention}, are employed to enhance model expressiveness and capture diverse token interactions. Multi-head attention involves projecting the input into multiple subspaces, each processed by an independent attention mechanism. The outputs are then concatenated and linearly transformed to produce the final attention output.

\subsection{Pooling Strategies}
The attention mechanism generates token-level embeddings that must be aggregated to form sentence-level representations. Pooling layers play a crucial role in this aggregation process, condensing token embeddings into unified sentence embeddings. We explore three primary pooling strategies: Mean, Max, and Weighted Sum pooling.

Given the token-level embeddings $X = \{x_1, x_2, \ldots, x_l\}$ produced by the attention mechanism, where $x_i \in \mathbb{R}^e$ and $l$ represents sequence length, we explore three primary pooling strategies:

\subsubsection{Mean Pooling}
\begin{equation}
    \text{Pool}_{\text{mean}}(X) = \frac{1}{l} \sum_{i=1}^l x_i
\end{equation}

Mean pooling provides a balanced representation by averaging token embeddings, offering a uniform perspective across all tokens.

\subsubsection{Max Pooling}
\begin{equation}
    \text{Pool}_{\text{max}}(X) = \max_{i=1}^l x_i
\end{equation}

Max pooling captures the most salient features by selecting the maximum values across token dimensions $e$, emphasizing distinctive embedding characteristics.

\subsubsection{Weighted Sum Pooling}
\begin{equation}
    \text{Pool}_{\text{ws}}(X) = \sum_{i=1}^l w_i x_i
\end{equation}
where $W = \{w_1, w_2, \ldots, w_l\}$ represents learnable weight parameters. This approach dynamically assigns importance to different tokens through learned weights.

Notably, mean and max pooling emerge as special cases of weighted sum pooling:
\begin{itemize}
    \item Mean pooling occurs when weights are uniform ($w_i = \frac{1}{l}$)
    \item Max pooling approximates when weights concentrate on the most significant token
\end{itemize}

While weighted sum pooling introduces additional parameters, the linear parameter increment remains negligible compared to the model's overall complexity.

\section{Experiments}
Our comprehensive evaluation investigates three pooling operations across BERT and GPT2 models for sentiment analysis, providing insights into their differential performance and practical implications.

\subsection{Datasets}
We employed the IMDB movie review sentiment analysis dataset, a benchmark corpus for binary sentiment classification. The dataset comprises 50,000 movie reviews meticulously balanced between positive and negative sentiments. Our data partition strategy follows a standard split:
\begin{itemize}
    \item Training set: 60\% (30,000 samples)
    \item Validation set: 10\% (5,000 samples)
    \item Testing set: 30\% (15,000 samples)
\end{itemize}

The deliberate equal distribution of positive and negative samples mitigates class imbalance, ensuring robust model training and evaluation. This balanced approach prevents potential biases that could arise from skewed class representations. Table \ref{tab:dataset} provides a summarization.

\begin{table}[htbp]
    \centering
    \caption{Dataset Summarization}
    \begin{tabular}{lccc}
        \toprule
        \multicolumn{1}{c}{\multirow{2}[0]{*}{IMDB}} & \multirow{2}[0]{*}{\#Samples} & \multicolumn{2}{c}{\multirow{2}[0]{*}{\#Categories}}              \\
                                                     &                               & \multicolumn{2}{c}{}                                              \\
        \midrule
        Train                                        & 30,000                        & 16,358 (P)                                           & 13,642 (N) \\
        Validation                                   & 5,000                         & 2,466 (P)                                            & 2,534 (N)  \\
        Test                                         & 15,000                        & 7,634 (P)                                            & 7,366 (N)  \\
        \bottomrule
    \end{tabular}%
    \label{tab:dataset}%
\end{table}%

\subsection{Experiment Settings}

We selected pretrained BERT-base and GPT2 as our baseline models, implementing a rigorous experimental protocol:

\subsubsection{Model Configuration}
\begin{itemize}
    \item Backbone learning rate: 1e-5
    \item Newly added parameters learning rate: 1e-3
    \item Remaining hyperparameters: Default configurations as specified in original model papers
\end{itemize}

\subsubsection{Evaluation Metrics}
We adopted a comprehensive suite of performance metrics to provide multifaceted model assessment:

\begin{itemize}
    \item Confusion Matrix: A tabular visualization depicting the model's prediction accuracy by categorizing outcomes into true positives, true negatives, false positives, and false negatives.
    \item Precision ($P = \frac{TP}{TP + FP}$): Quantifies the proportion of correctly predicted positive instances among all positive predictions, measuring the model's exactness.
    \item Recall ($R = \frac{TP}{TP + FN}$): Represents the proportion of actual positive instances correctly identified, assessing the model's completeness.
    \item F1 Score ($F1 = 2 \times \frac{P \times R}{P + R}$): Harmonic mean of precision and recall, providing a balanced metric that simultaneously considers both false positives and false negatives.
\end{itemize}

\subsection{Results}

\subsubsection{Confusion Matrix Analysis}

\begin{figure}[htbp]
    \centering
    \begin{subfigure}[b]{0.32\linewidth}
        \centering
        \includegraphics[width=\linewidth]{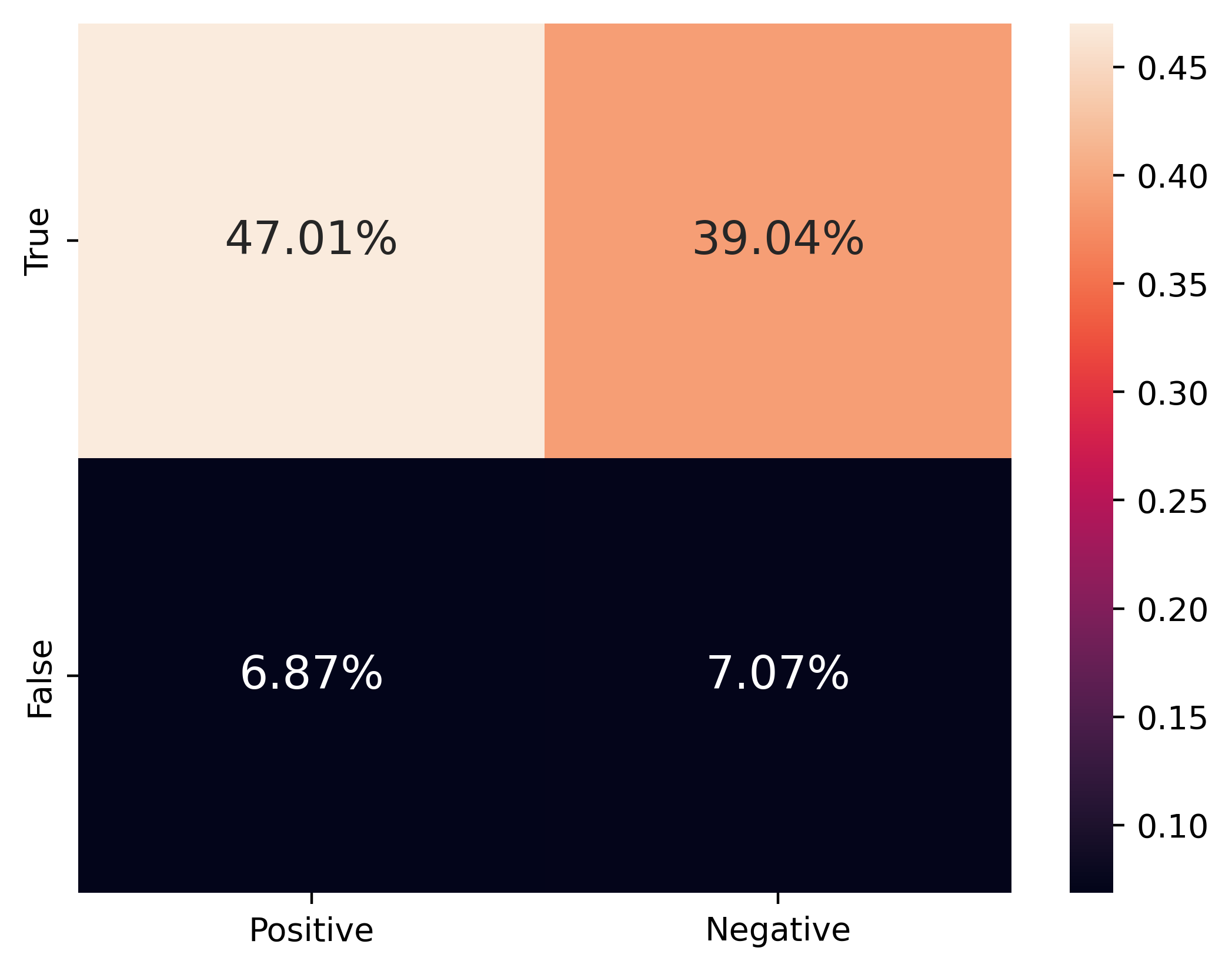}
        \caption{Mean}
        \label{fig1:subfig1}
    \end{subfigure}
    \hfill
    \begin{subfigure}[b]{0.32\linewidth}
        \centering
        \includegraphics[width=\linewidth]{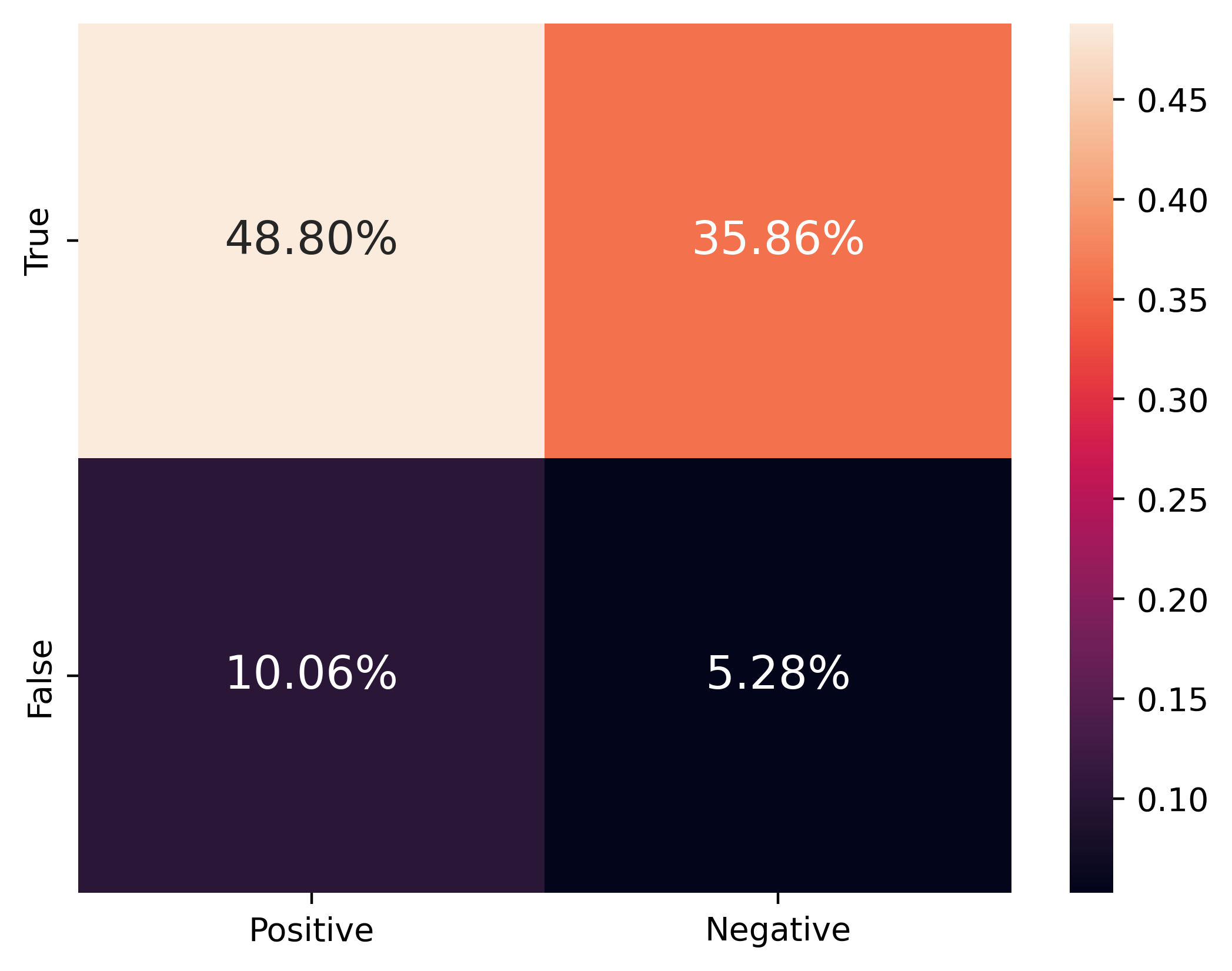}
        \caption{Max}
        \label{fig1:subfig2}
    \end{subfigure}
    \hfill
    \begin{subfigure}[b]{0.32\linewidth}
        \centering
        \includegraphics[width=\linewidth]{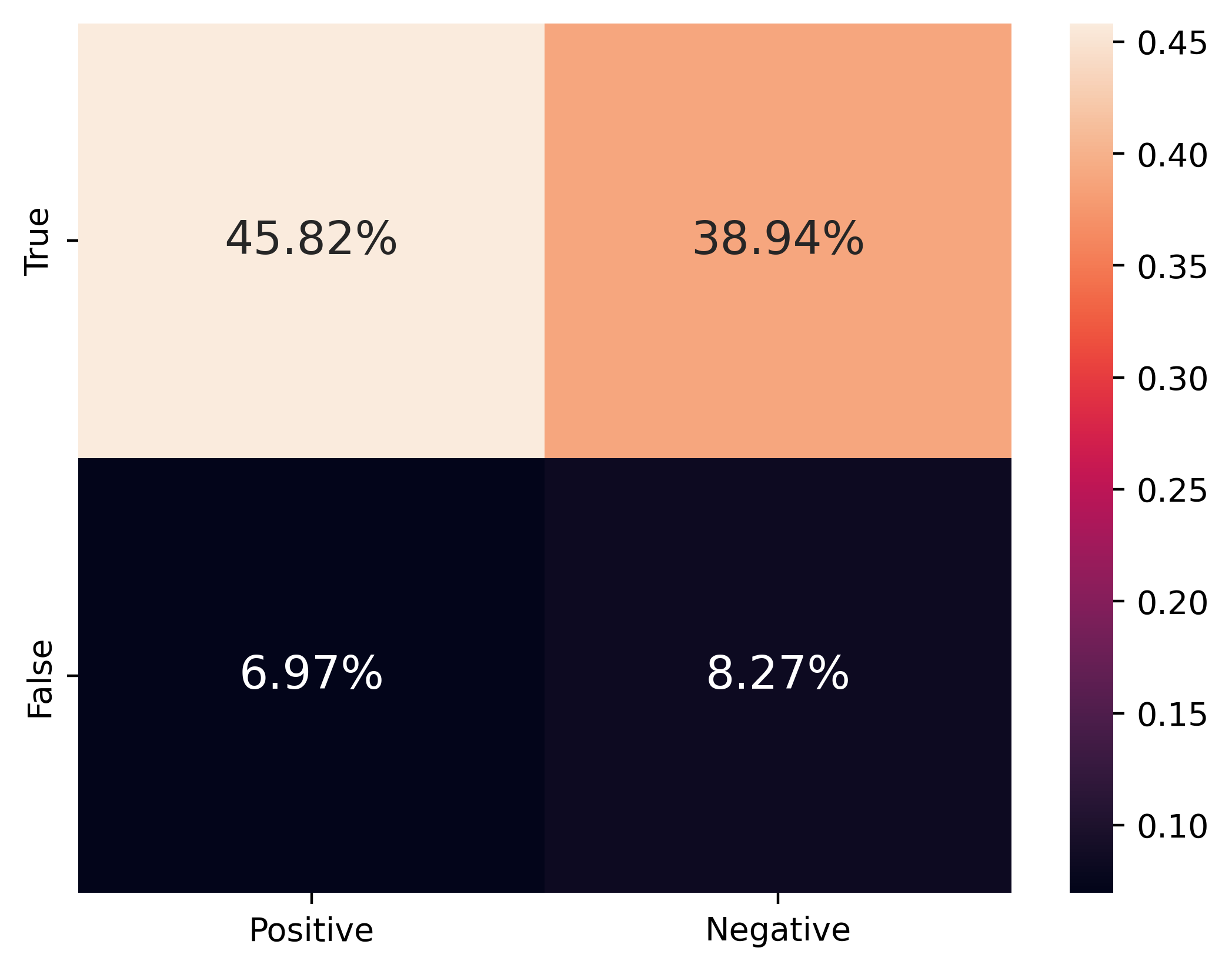}
        \caption{Weighted Sum}
        \label{fig1:subfig3}
    \end{subfigure}
    \caption{Confusion Matrix of Bert with Different Pooling Strategies}
    \label{fig1:mainfig}
\end{figure}

\begin{figure}[htbp]
    \centering
    \begin{subfigure}[b]{0.32\linewidth}
        \centering
        \includegraphics[width=\linewidth]{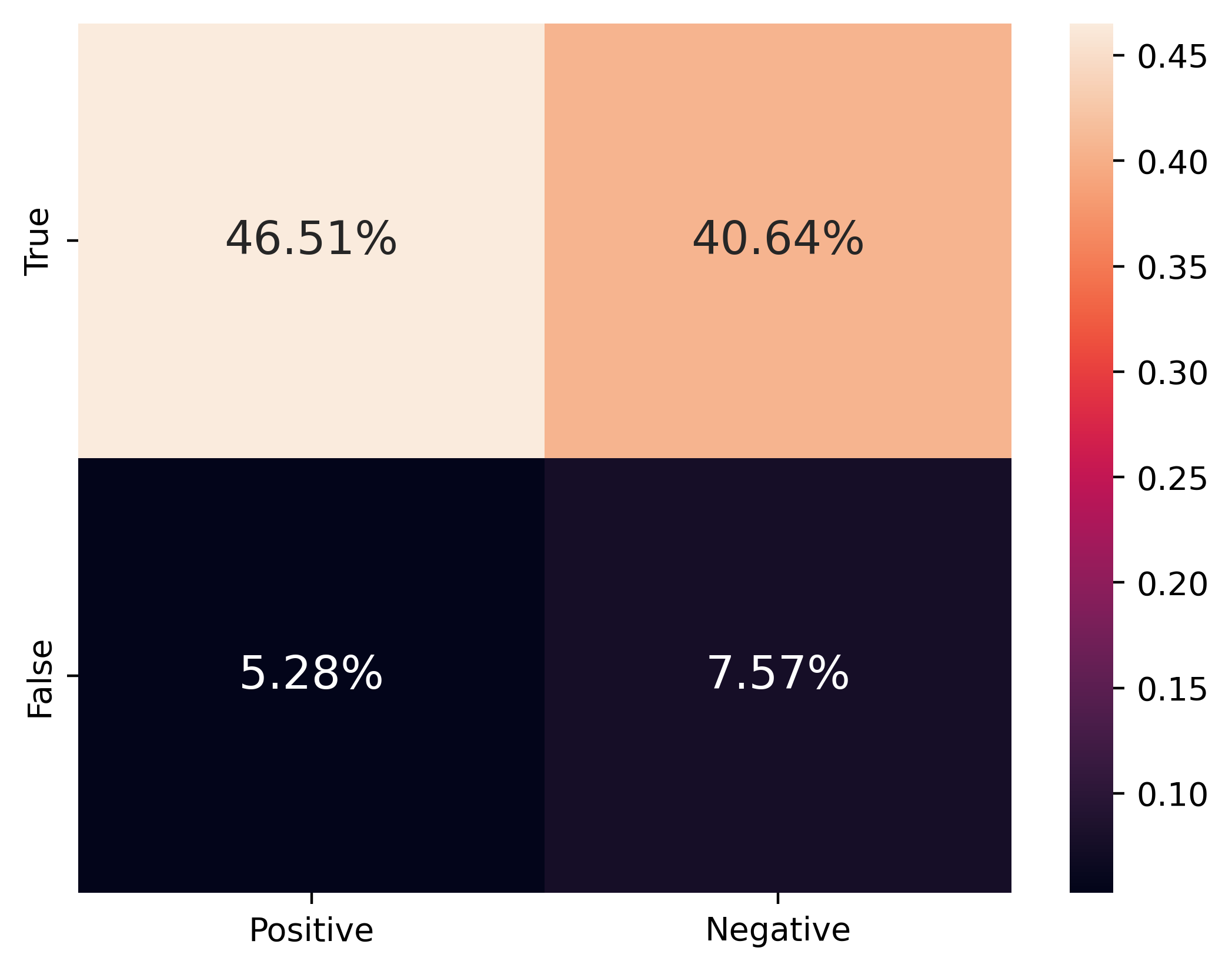}
        \caption{Mean}
        \label{fig2:subfig1}
    \end{subfigure}
    \hfill
    \begin{subfigure}[b]{0.32\linewidth}
        \centering
        \includegraphics[width=\linewidth]{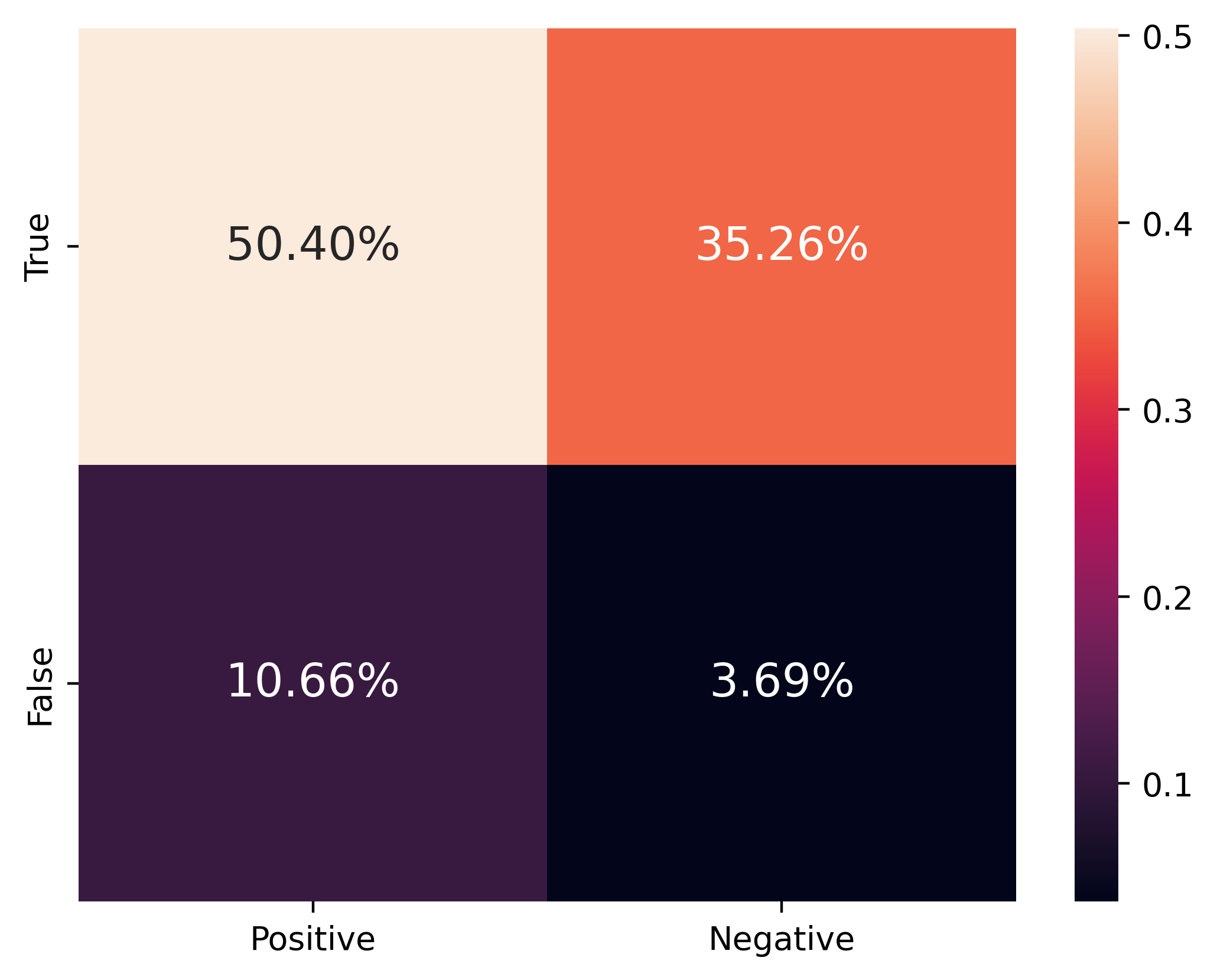}
        \caption{Max}
        \label{fig2:subfig2}
    \end{subfigure}
    \hfill
    \begin{subfigure}[b]{0.33\linewidth}
        \centering
        \includegraphics[width=\linewidth]{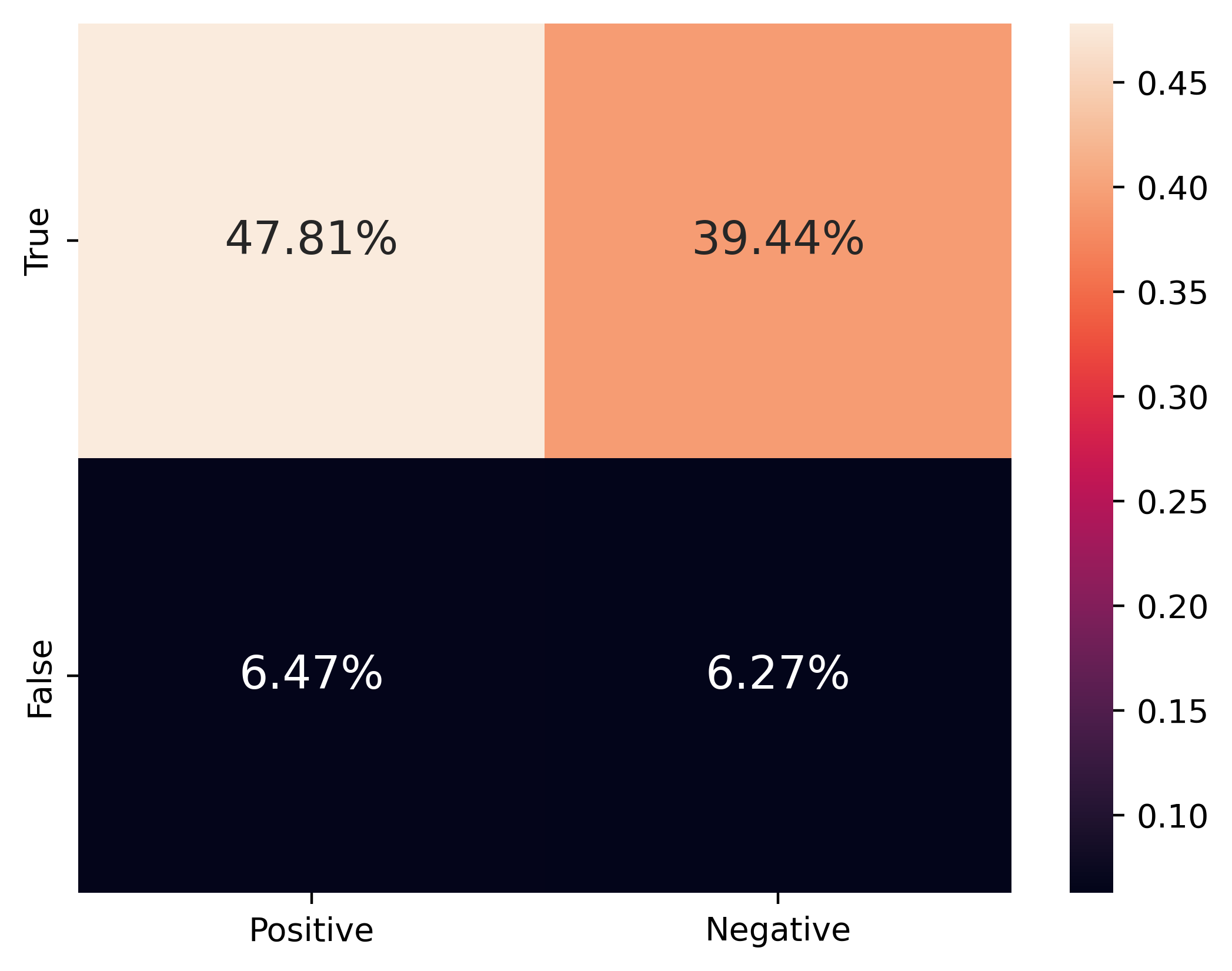}
        \caption{Weighted Sum}
        \label{fig2:subfig3}
    \end{subfigure}
    \caption{Confusion Matrix of GPT2 with Different Pooling Strategies}
    \label{fig2:mainfig}
\end{figure}

As shown in Figure \ref{fig1:mainfig}-\ref{fig2:mainfig}, both BERT and GPT2 demonstrated comparable performance across pooling mechanisms, with an aggregate correct prediction rate of approximately 86.07\%. 

For BERT, Max pooling exhibited distinctive characteristics:
\begin{itemize}
    \item Highest True Positive rate
    \item Lowest True Negative rate
    \item Potentially attributed to the max operation's tendency to select most prominent features
\end{itemize}

Mean and Weighted Sum pooling displayed more balanced performance, leveraging comprehensive token consideration. Similar observations can be drawn from GPT2.

\subsubsection{Performance Metrics Evaluation}

\begin{figure}[htbp]
    \centering
    \begin{subfigure}[b]{0.49\linewidth}
        \centering
        \includegraphics[width=\linewidth]{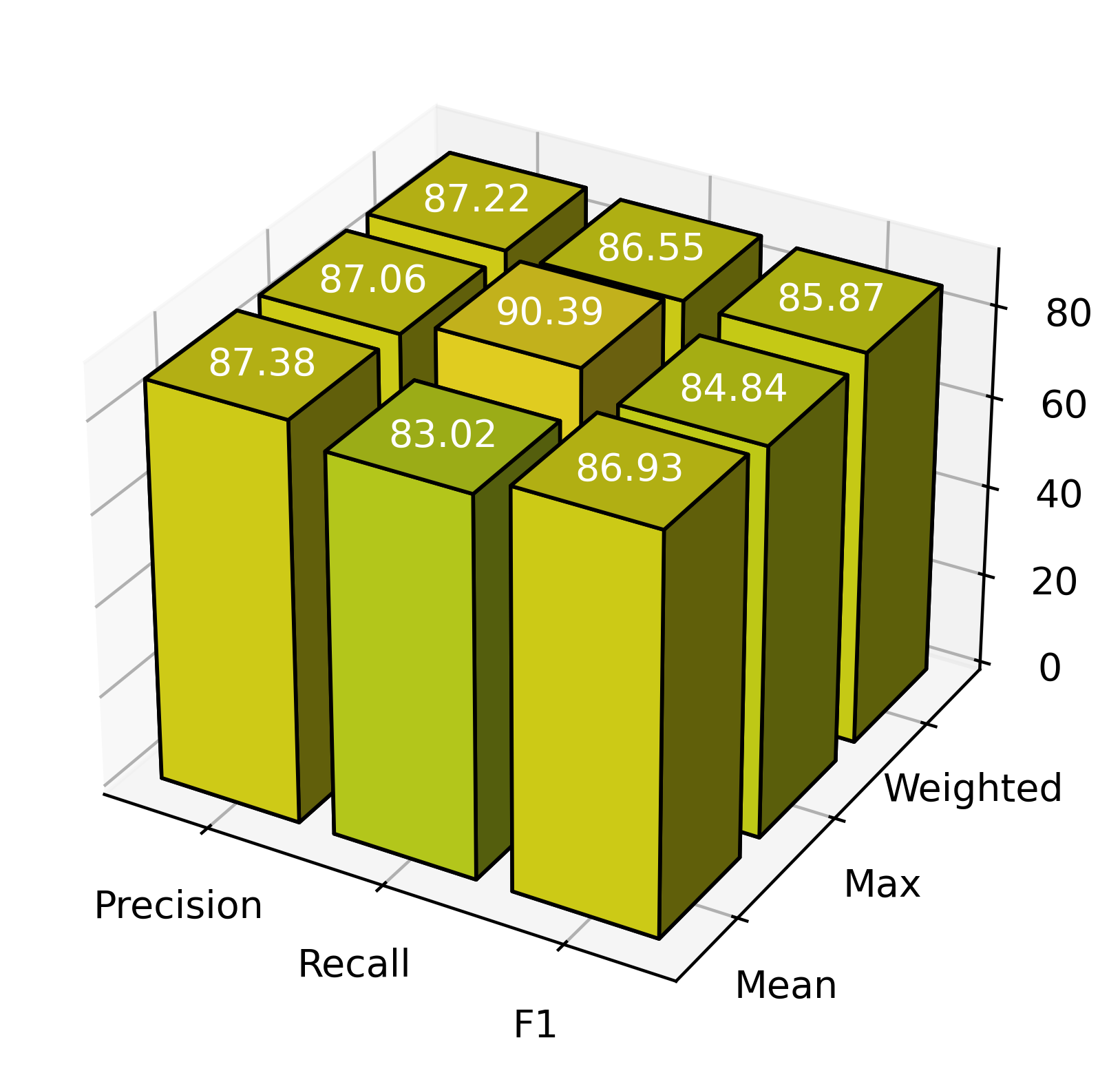}
        \caption{Scores of Bert (\%)}
        \label{fig3:subfig1}
    \end{subfigure}
    \hfill
    \begin{subfigure}[b]{0.49\linewidth}
        \centering
        \includegraphics[width=\linewidth]{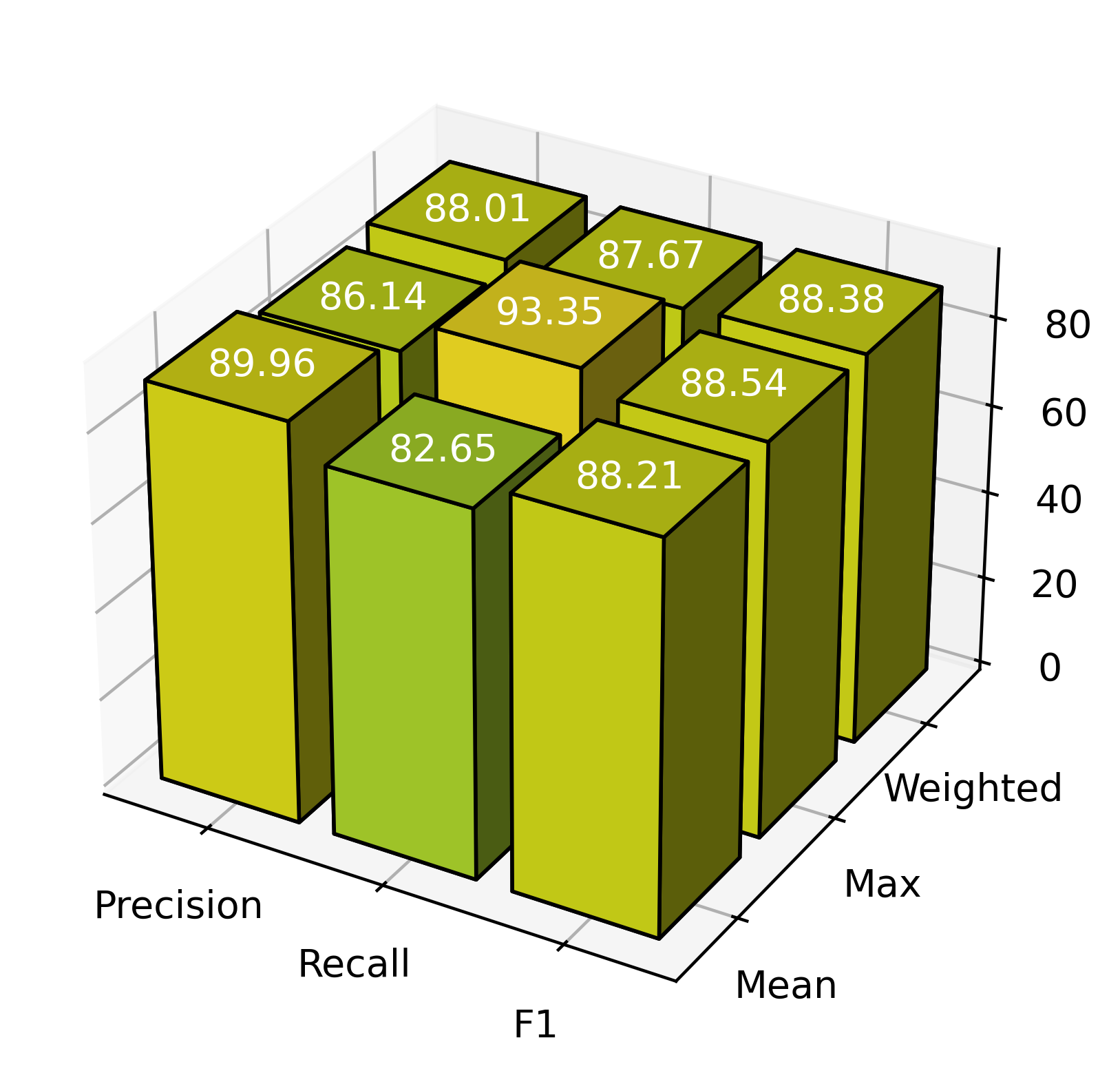}
        \caption{Scores of GPT2 (\%)}
        \label{fig3:subfig2}
    \end{subfigure}
    \caption{Precision, Recall, F1 Score of Bert and GPT2 with Different Pooling Strategies}
    \label{fig3:mainfig}
\end{figure}

Precision, Recall, and F1 score analyses depicted in Figure \ref{fig3:mainfig} revealed nuanced insights:
\begin{itemize}
    \item BERT achieved peak performance (87.22\%) with Mean pooling
    \item GPT2 optimized performance (88.38\%) using Weighted Sum pooling
\end{itemize}

These results underscore the model-specific nature of pooling mechanism effectiveness. Key observations include:
\begin{itemize}
    \item Mean pooling offers computational efficiency and balanced representation
    \item Weighted Sum provides enhanced flexibility for complex tasks
    \item Model architecture significantly influences pooling mechanism performance
\end{itemize}

\subsubsection{Practical Recommendations}
\begin{itemize}
    \item For computational constraints: Utilize Mean pooling
    \item For complex, adaptable applications: Employ Weighted Sum
    \item When maximum positive detection is crucial: Consider Max pooling
\end{itemize}

The findings emphasize that pooling mechanism selection should be tailored to specific model architectures, downstream tasks, and computational resources.

\section{Conclusion}
This paper provides a systematic evaluation of pooling mechanisms -- Mean, Max, and Weighted Sum on two leading LLM families, BERT and GPT, with a focus on sentiment analysis as a representative sentence-level task. Our findings highlight the distinctive contributions of each pooling strategy: Mean pooling excels in general scenarios by providing stable and robust embeddings, Max pooling emphasizes salient features but may overfit to extremes, and Weighted Sum pooling offers flexibility but requires careful optimization. These results emphasize the importance of tailoring pooling mechanisms to align with task-specific requirements and model architectures.

By delving into the interplay between pooling strategies and LLM performance, this research fosters a deeper understanding of the design choices in pooling layers and their impact on downstream applications. Practical recommendations drawn from our analysis aim to guide researchers and practitioners in selecting appropriate pooling methods for their needs. Looking ahead, extending this investigation to additional architectures, tasks, and pooling variations holds promise for further refining the use of LLMs in NLP.

\end{document}